\documentclass{article}

\usepackage[preprint]{neurips_2022}

\usepackage{latexsym}
\usepackage{mathtools}
\usepackage{booktabs}
\usepackage{array}
\usepackage{tabularray}
\UseTblrLibrary{booktabs}
\usepackage{graphicx}
\usepackage{tikz}
\usepackage{pgfplots}
\pgfplotsset{compat=1.18}
\usepackage{float}
\usepackage{wrapfig}
\usepackage{needspace}
\usepackage{placeins}
\usepackage{enumitem}
\usepackage{times}
\usepackage{subfigure}
\usepackage{algorithm}
\usepackage{algorithmic}
\usepackage{sidecap}
\usepackage{hyperref}
\usepackage{amssymb}
\usepackage{xcolor}
\usepackage[normalem]{ulem}

\definecolor{shadecolor}{gray}{0.95}

\begin{document}

\title{Unconstrained CTI-REALM with Open-Weights}

\author{
  Sahil Garg\thanks{Author names are in descending order of contribution, not an indication of seniority.},
  \ 
  Ruhi Gottumukkala, Vishal Agarwal, Shubham Vishwakarma\\
  Averlon \\
  Redmond, WA \\
  Correspondence: sahil.garg@averlon.io, sahil.garg.cs@gmail.com
}

\maketitle

\begin{abstract}
Organisations that want language-model agents for security work face a choice between
open-weight models, which they can run and audit themselves, and far more expensive closed
frontier models. We study that choice on CTI-REALM\footnote{A.~Chakraborty, S.~Ho, A.~Cook and M.~Mel\'{e}ndez, \emph{CTI-REALM: Benchmark to evaluate agent performance on security detection rule generation capabilities}, 2026, \href{https://arxiv.org/abs/2603.13517}{arXiv:2603.13517}.}, a benchmark in which an agent turns a
threat-intelligence report into a working detection by querying a live telemetry store. We
evaluate twenty-six models, nineteen open-weight and seven closed, under a common harness,
scorer and panel of LLM judges. Our main finding is that an open-weight model can match the
strongest closed models at less than a tenth of their cost, provided it is given enough
turns. The benchmark's published limit on agent messages is a confound: a task that exhausts
the limit scores zero on every checkpoint, including those it had already passed, so results
at that limit partly measure which models finish in time. At the published limit the leading
open-weight and closed models cannot be distinguished. With the limit relaxed, the models it
was truncating recover, and the highest rewards belong to open-weight models. Among the
leading models, price is a poor guide to capability: models with similar rewards differ in
cost by more than an order of magnitude. Because the benchmark's checkpoint scores largely
record what an agent did or mentioned, we also introduce stage-level metrics that score what
the agent submits, by executing its detection queries in the telemetry store and judging its
outputs against the answer key. We deliberately do not combine them into a new overall reward,
which would be one more weighting among many defensible ones. Instead they show that different
models lead different stages, which suggests routing each stage of the workflow to the model
that performs it best rather than choosing a single model by its overall score. We also report
how many tokens each model generates, and find that the share of output spent on thinking does
not track the reward.
\end{abstract}

\section{Introduction}
\label{sec:intro}

Security operations teams find intrusions by running detection rules over the logs their
systems emit. Those logs are gathered in a security information and event management system
(SIEM): a central store that collects telemetry from endpoints, servers, cloud services and
identity providers, evaluates detection rules against it continuously, and raises an alert
for an analyst whenever a rule matches. A SIEM is only as good as its rules, and writing them
is the job of a detection engineer. Handed a threat-intelligence report on a new intrusion
campaign, the engineer reads it and decides which adversary behaviours it describes, works
out which of several hundred telemetry tables would have recorded them, queries that
telemetry to check that the evidence is there, and only then writes the detection. Each step
depends on the one before it, and only the last step produces a rule that is deployed.

This is the kind of multi-step work that organisations now want agentic language models to
take on, and the deployment question is usually posed as a choice between open-weight and
frontier closed models: can a model whose weights can be run and audited in-house do the work of a far more
expensive proprietary one? CTI-REALM \citep{chakraborty2026ctirealm} is a natural benchmark on which to
ask this question. It follows the engineer's workflow in five checkpoints, from reading the report to
submitting a detection, over intrusions emulated on Linux hosts, cloud control planes and
Kubernetes, and the agent must itself query a live log store in the Kusto Query Language (KQL), the query language
of Microsoft's Sentinel SIEM.

\begin{table}[t]
\centering
\small
\begin{tabular}{@{}ll@{}}
\toprule
Developer & Models evaluated \\
\midrule
\multicolumn{2}{@{}l}{\emph{Open-weight (19 models)}} \\
Zhipu AI & GLM-5.3-flash, GLM-5.3, GLM-5.2 \\
DeepSeek & DeepSeek-v4.1-flash, DeepSeek-v4-flash, DeepSeek-v4-pro \\
Alibaba & Qwen3.8-2.4T, Qwen3.8-27B, Qwen3.6-35B-A3B \\
Moonshot AI & Kimi-K3 \\
MiniMax & MiniMax-M3 \\
Google DeepMind & Gemma-4-31B, Gemma-4-26B-A4B \\
NVIDIA & Nemotron-3-Ultra \\
Thinking Machines Lab & Inkling, Inkling-Small \\
Mistral AI & Mistral-Large-4 \\
OpenAI & GPT-OSS-120B, GPT-OSS-20B \\
\midrule
\multicolumn{2}{@{}l}{\emph{Closed (7 models)}} \\
OpenAI & GPT-5.6-sol, GPT-5.6-terra, GPT-5.6-luna  \\
Anthropic & Claude-Opus-5, Claude-Sonnet-5, Claude-Haiku-4.5  \\
xAI & Grok-4.6  \\
\bottomrule
\end{tabular}
\par\vspace{6pt}
\caption{The twenty-six models evaluated, grouped by whether their weights are published.
The closed models are available only through their developers' APIs. Mistral-Large-4 is counted as open-weight: its developer released it on 6~October~2026 with its weights announced for publication by the end of that month.}
\label{tab:models}
\end{table}

We find that the answer depends on a setting that is easy to mistake for part of the
benchmark. CTI-REALM caps every task at a fixed number of agent messages, and the cap is
hard: a task that exhausts it never submits, and so scores zero on every checkpoint,
including the ones it had already passed. At the published limit the strongest open-weight
and closed models cannot be distinguished, and part of what is measured there is which models
finish within the limit. With the limit relaxed, the open-weight models it was truncating
recover, and \emph{an open-weight model matches the strongest closed models at a small
fraction of their cost} (Figure~\ref{fig:cost_accuracy_b300}).

With the limit relaxed, the two highest rewards belong to open-weight models. The differences
among the leading models are small, and we do not claim an ordering among them
(Section~\ref{sec:overall}). Our claim is parity: the best open-weight models match the best
closed models on this benchmark, and one of them does so at a cost low enough that the extra
turns add little to it. A second observation concerns price: among the leading models the
cost of a run varies by more than an order of magnitude, so on this task price says little
about how well a model does the work.

A third observation concerns the benchmark's stage scores, which mostly record what an agent did
or mentioned rather than what it submitted. For four of the five stages we therefore introduce a
metric that scores the submission itself against the benchmark's answer key, by executing the
submitted detection query in the telemetry store or by asking LLM judges to match the submission
to the key. Where the benchmark's score measures something else, such as the agent's exploration,
these metrics complement it; where it measures the same quantity poorly, they correct it. We do not combine these metrics into a new overall reward: the weighting would be one
more choice among many defensible ones, and a single score would hide what the metrics show,
namely that no model leads every stage. The best system for this workflow is therefore not
necessarily one model chosen by its overall ranking, but a route that hands each stage to the
model that performs it best.

\subsection{Contributions}
\label{sec:contributions}

\begin{enumerate}
\item \emph{Open-weight models match the strongest closed models, one of them at a small fraction of the cost.}
Given enough turns to finish, the two best open-weight models score level with or above the
two strongest closed models we measured, and one of them does so at less than a tenth of
their cost. Section~\ref{sec:overall} gives the rewards, the costs and a paired bootstrap
interval for the gap between the leading models.

\item \emph{Separation of the message budget from model capability.} A task that
exhausts the limit scores zero on every checkpoint, including ones it had already passed,
so the published setting measures termination as well as capability. We therefore report
every model at an unconstrained budget, at which no reported sample is truncated
(Section~\ref{sec:overall}).

\item \emph{Weak relation between price and capability among the leading models.} Among the models that score close to the
leader, the cost of a run varies by more than an order of magnitude
(Section~\ref{sec:overall}).

\item \emph{An open-against-closed evaluation under a common harness, scorer and judge panel.}
We evaluate twenty-six models, nineteen open-weight and seven closed (Table~\ref{tab:models}), and
report each on each of the five checkpoints and on the overall reward
(Sections~\ref{sec:setup} and~\ref{sec:overall}).

\item \emph{Stage-level metrics grounded in the answer key.} For four of the five stages we
introduce metrics that score what the agent submits rather than what it did or mentioned:
judge-matched ATT\&CK mapping, the tables the submitted query reads against those that record
the attack in the store, whether that query was tested before submission, and the detection itself, executed in the telemetry store and judged
against the answer key. They complement the benchmark's stage scores, or correct them where they
mismeasure. We do not define a new overall reward. The metrics show that different
models lead different stages, which argues for routing each stage to a different model rather
than choosing one model by its overall reward, and building them exposes tasks whose answer key
does not match the benchmark's own data (Section~\ref{sec:stages}).

\item \emph{Output token usage.} We count the output tokens behind every reported
reward, thinking included and checked against the visible text, and relate the volume and
the thinking share to the reward (Section~\ref{sec:tokens}).
\end{enumerate}

\paragraph{Organisation.} Section~\ref{sec:setup} describes the benchmark and how
we ran it: the five checkpoints and the reward, the judge panel, the message budget and how
cost is counted. A reader who wants only the outcome can skip it and go directly to
Section~\ref{sec:overall}, which gives the results: reward against cost at the unconstrained
budget (Figure~\ref{fig:cost_accuracy_b300}) and at the published limit
(Figure~\ref{fig:cost_accuracy_b70}), and the comparison between the leading models.
Section~\ref{sec:tokens} reports how many tokens a run generates and how much of that is
thinking (Figure~\ref{fig:tokens}). Section~\ref{sec:stages} introduces our own metrics for C1 to C4
beside the benchmark's own score and diagnostics for every checkpoint (Tables~\ref{tab:stage_c0}--\ref{tab:stage_c4}) and the answer-key defects they expose. Section~\ref{sec:conclusion} concludes, and
Section~\ref{sec:overall} also lists, in Table~\ref{tab:cti_realm_bunlim}, the benchmark's scores of every model on
every checkpoint.

\providecommand{\rankone}[1]{{\color[HTML]{0A46D8}\normalsize\bfseries #1}}
\providecommand{\ranktwo}[1]{{\color[HTML]{4169E1}\small\bfseries #1}}
\providecommand{\rankthree}[1]{{\color[HTML]{8442C4}\small #1}}
\providecommand{\subcell}[1]{{\footnotesize\color[HTML]{6B6478}#1}}
\definecolor{rowtint}{HTML}{F1EDF7}
\definecolor{accentblue}{HTML}{0A46D8}

\section{Evaluation Setup}
\label{sec:setup}

All results are on CTI-REALM \citep{chakraborty2026ctirealm}, over the same fixed set of \emph{50 agentic tasks} spanning Linux, cloud and Kubernetes (Azure Kubernetes Service, AKS) emulations. Each task runs a five-checkpoint detection-engineering workflow: \emph{C0} retrieve and read the threat-intelligence report, \emph{C1} map it to techniques of MITRE ATT\&CK, a public catalogue of adversary techniques, \emph{C2} discover which telemetry tables hold the evidence, \emph{C3} execute queries against the store, and \emph{C4} produce a KQL detection and a Sigma rule, a detection written in an open rule format that is portable across SIEMs.

We run the benchmark's public implementation, the \texttt{cti\_realm\_50} task of \texttt{inspect\_evals}, with its tasks, answer keys and judge prompts. The one change is the judge: the judged checkpoints are graded by the panel described below in place of the implementation's default judge, GPT-5-mini, so our absolute scores are not directly comparable with those reported by \citet{chakraborty2026ctirealm}.

\paragraph{Reward.} For one task, the benchmark's overall reward is
\begin{equation}
R \;=\; \frac{1}{10}\sum_{j=0}^{4} w_j\, s_j ,\qquad
(w_0,\ldots,w_4) \;=\; (1.25,\; 0.75,\; 1.00,\; 0.50,\; 6.50).
\label{eq:reward}
\end{equation}
Here, the checkpoint scores and their weights satisfy
\begin{equation*}
s_j \in [0,1], \qquad \sum_{j=0}^{4} w_j = 10 .
\end{equation*}
Table~\ref{tab:cti_realm_bunlim} reports each checkpoint score on its own 0--1 scale rather than as its weighted contribution to the reward, so that it can be read against the checkpoint's own ceiling; \emph{overall} is the reward. Note that the weight of C4 alone is 6.5 of the ten points, so the final detection carries more weight than the four preceding checkpoints combined. C4 itself has two parts: the F1, against the answer key, of the result rows the agent reports in its answer for its submitted KQL query (5.0 of its 6.5 points), and the judged quality of the Sigma rule (1.5 points):
\begin{equation*}
s_4 \;=\; \frac{5.0\,\mathrm{F1}_{\text{query}} \;+\; 1.5\,q_{\text{Sigma}}}{6.5} .
\end{equation*}
Here, the numerator combines the query F1 and the judged quality of the Sigma rule. The C4 column of Table~\ref{tab:cti_realm_bunlim} reports the query F1; the Sigma part enters \emph{overall} but has no column.

A \emph{sample} is one model's attempt at one task, and it is \emph{truncated} when it exhausts its message budget. A truncated sample enters Eq.~\eqref{eq:reward} with \emph{every} checkpoint score at zero, not merely the last: it never reaches \texttt{submit()}, and the scorer sees no submission to grade, so checkpoints the agent had already passed score zero as well. Truncation therefore acts on the reward through all five terms at once, which is why the number of truncated samples is the first thing to check when a reward at the published limit looks low.

We report the sample mean of the reward over the 50 tasks,
\begin{equation*}
\hat{R} \;=\; \frac{1}{50}\sum_{\text{task } i} R_i .
\end{equation*}
This mean estimates the reward this workflow would obtain on the population of intrusions the 50 tasks are drawn from, and the estimate and that population value should not be conflated: Section~\ref{sec:overall} gives paired bootstrap intervals for the gaps between the leading models.

\paragraph{Judging.} C0 and the Sigma part of C4 (1.5 of its 6.5 points) are graded by LLM judges; the other checkpoints are scored against the answer key, so 72.5\% of the reward weight involves no judge. The benchmark's implementation grades each judged answer with a single call to one judge, GPT-5-mini, at temperature zero. One call is a noisy estimate of a judgement, and temperature zero is not recommended for current reasoning models; several, GPT-5-mini among them, no longer accept it. We therefore grade each judged answer with nine trials, three from each of three judges, Gemma-4-31B with its reasoning enabled, DeepSeek-v4-flash and GLM-5.3-flash, each otherwise at its default sampling and reasoning settings and allowed replies of up to 32,000 tokens rather than the implementation's 5,000, which a reasoning judge can exceed, and average within each judge before averaging across the three, so that every judge carries equal weight:
\begin{equation}
\hat{s}_j \;=\; \frac{1}{|\mathcal{J}|}\sum_{k \in \mathcal{J}} \frac{1}{T}\sum_{t=1}^{T} s_j^{(k,t)} ,\qquad |\mathcal{J}| = 3,\; T = 3 ,
\label{eq:panel}
\end{equation}
where the outer sum runs over the three judges and the inner one over each judge's three trials. A judge reply that carries no verdict is asked again, up to three times.

\paragraph{Judges that are also evaluated.} Three of the 26 models, DeepSeek-v4-flash, Gemma-4-31B and GLM-5.3-flash, are also judges on the panel and are graded by it like every other model. GLM-5.3-flash, the model with the highest reward, receives the same reward to two decimals whether or not it is among its own judges.

\paragraph{Thinking.} Every model runs at its default reasoning setting, with two exceptions. Gemma-4-31B and Gemma-4-26B-A4B have thinking off by default, which would compare models that cannot think with models that can, so we run both with thinking enabled. With thinking off, they score 0.39 and 0.35 at the published limit, against 0.52 and 0.47 with thinking on. None of Gemma-4-31B's 50 samples reached the cap, so its unconstrained result is carried under the rule stated next. One of Gemma-4-26B-A4B's samples reached the cap at the published limit and was re-run at 300, the splice described below. Claude-Haiku-4.5 does not think at its default setting and is the one model evaluated without thinking.

\paragraph{Message budget.} Each task is capped at a number of agent messages, its \emph{message budget}, and a sample that exhausts the budget has reached the \emph{cap}. We report two settings: the \emph{published limit} of 70 and an \emph{unconstrained budget} of 300. Where no sample of a model reached the cap at a lower budget, a higher budget cannot change its result. Such a result is \emph{carried}: it is drawn with a hollow marker in Figure~\ref{fig:cost_accuracy_b300} and is not a run at the budget shown. No budget is called ``unlimited'': trajectory length is stochastic, and GLM-5.3-flash, which ran two samples to a 250 cap, finished those same two in 91 and 60 messages when given 500.

\paragraph{Composition of the unconstrained results.} Of the 26 models, 14 have samples run at budget 300 itself, 4 of them all 50 samples and 10 only the samples that had reached a lower cap, and 12 have a result carried from a lower budget under the rule above. Every carried result comes from a run with \emph{zero} samples at the cap, so a higher budget could not have changed it, but these results are not runs at 300, and Figure~\ref{fig:cost_accuracy_b300} draws them with hollow markers. Of the 7 closed models, Claude-Opus-5 and Grok-4.6 have samples run at 300.

Of those 14, 4 are complete 50-sample runs at budget 300 and 10 are \emph{spliced}: GLM-5.3-flash, Kimi-K3, Grok-4.6, Claude-Opus-5, Qwen3.8-2.4T, DeepSeek-v4-pro, Nemotron-3-Ultra, Gemma-4-26B-A4B, GPT-OSS-120B, GPT-OSS-20B. A \emph{splice} applies the carry rule per sample rather than per model: the model was run at a lower budget, the samples that finished within that budget were kept unchanged because a higher budget cannot alter them, and only the samples that reached the cap were re-run at 300. The splice rests on the same argument as the carry rule, applied one sample at a time. We name the spliced models because the distinction is invisible in Table~\ref{tab:cti_realm_bunlim} and in Figure~\ref{fig:cost_accuracy_b300}: a spliced model is not marked as carried, since part of its row is a run at 300.

\paragraph{Cost.} The \emph{cost} of a run is the cost of its model calls over the 50 tasks behind the reported reward, at list prices and never at a discounted or blended rate. Table~\ref{tab:cti_realm_bunlim} shows it in whole dollars, or in cents below a dollar, and the text follows the same convention; a ratio between two costs is computed from the unrounded costs. No cache-read rate is published for the GPT-5.6 family, so we price its cache reads at zero; these runs are 81--82\% cache reads, and their costs are therefore lower bounds rather than estimates.

\section{Overall Results}
\label{sec:overall}

\begin{figure}[!ht]
\centering
\includegraphics[width=\linewidth]{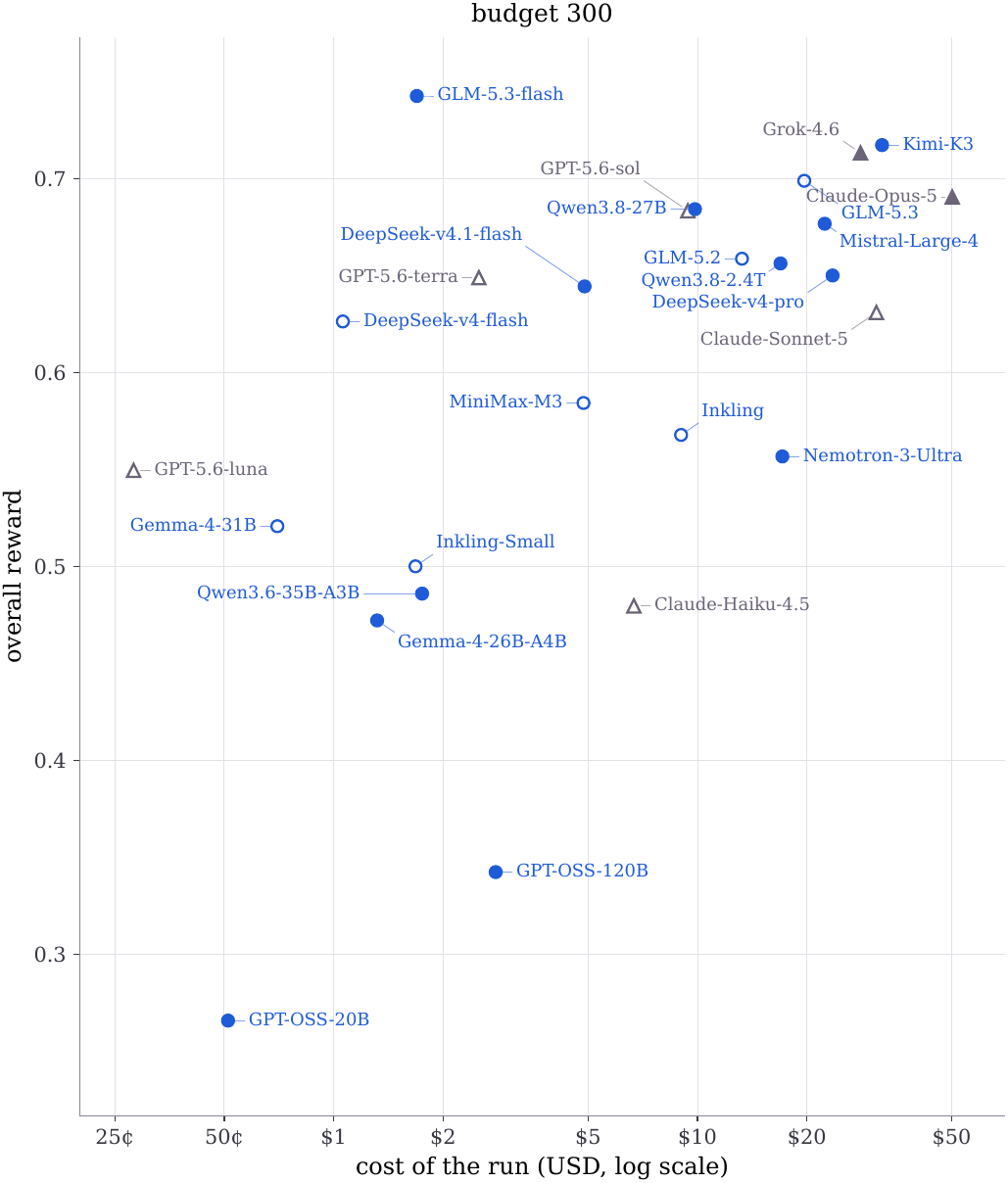}
\caption{Overall reward against the cost of the run at the unconstrained budget of 300 agent messages. This figure and Figure~\ref{fig:cost_accuracy_b70} share axes so that they can be compared. Open-weight models are drawn as blue circles and closed models as grey triangles. A hollow marker is a carried result: no sample of that model reached the cap at a lower budget, so the point is its result at that budget and not a run at 300. The highest reward belongs to GLM-5.3-flash, an open-weight model, at \$2. Note that the least expensive open-weight run shown is GPT-OSS-20B at 51\textcent{}.}
\label{fig:cost_accuracy_b300}
\end{figure}

\begin{figure}[!ht]
\centering
\includegraphics[width=\linewidth]{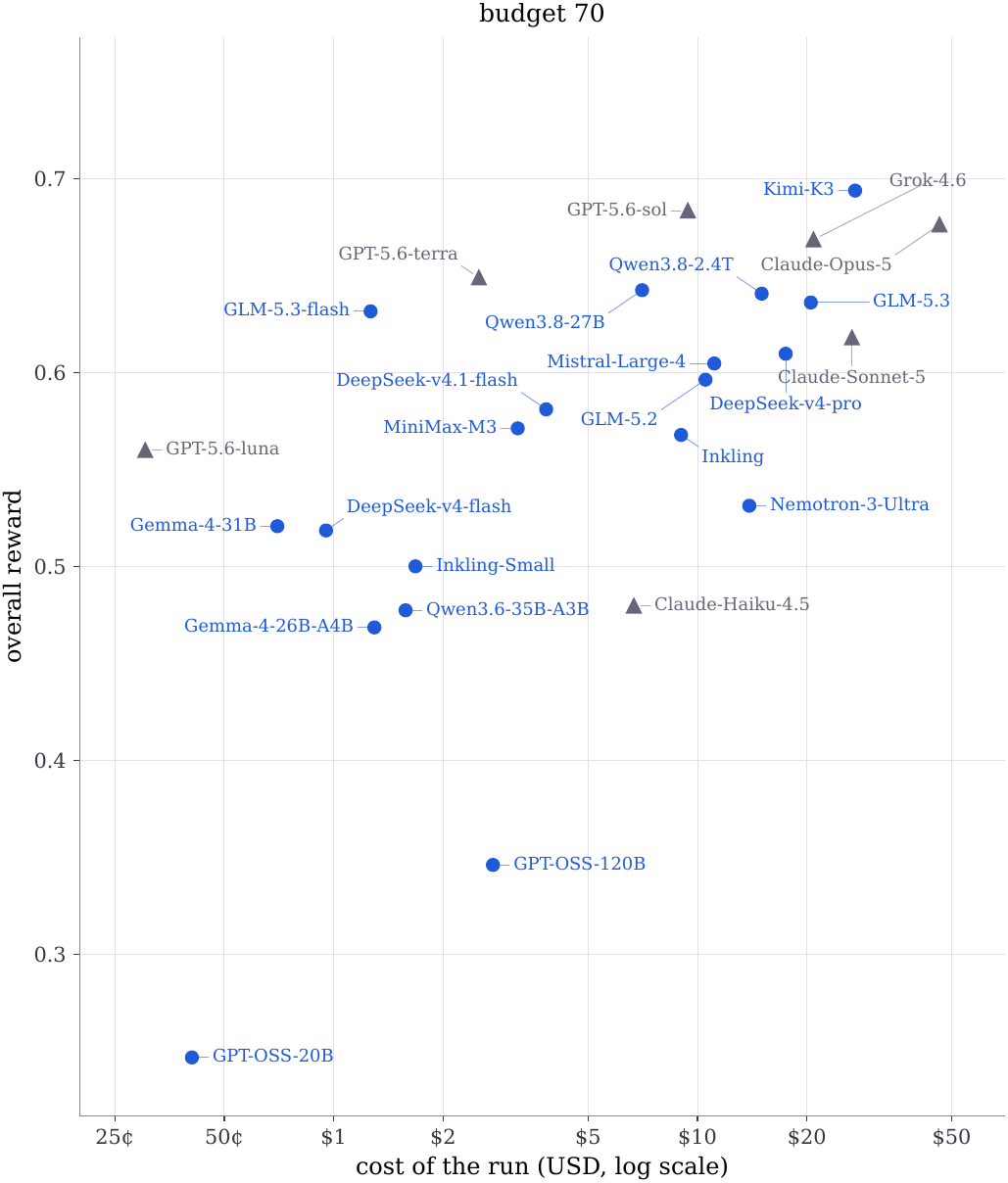}
\caption{Overall reward against the cost of the run at the benchmark's published limit of 70 agent messages. This figure and Figure~\ref{fig:cost_accuracy_b300} share axes so that they can be compared. Open-weight models are drawn as blue circles and closed models as grey triangles. The highest reward belongs to Kimi-K3, an open-weight model, at \$27. Note that the least expensive open-weight run shown is GPT-OSS-20B at 41\textcent{}.}
\label{fig:cost_accuracy_b70}
\end{figure}

Figures~\ref{fig:cost_accuracy_b300} and~\ref{fig:cost_accuracy_b70} plot the overall reward of all 26 models against the cost of the run, first at the unconstrained budget of 300 messages and then at the published limit of 70. Table~\ref{tab:cti_realm_bunlim} lists the numbers behind the first figure, together with the benchmark's five checkpoint scores. We observe that the reward varies far less across the models than the cost of a run does. At the unconstrained budget the 26 rewards span 0.27 to 0.74, a factor of about 2.8 from the lowest reward to the highest, while the cost of a fully priced run ranges from 51\textcent{} (GPT-OSS-20B) to \$50 (Claude-Opus-5), a factor of about 98; GPT-5.6-luna's 28\textcent{} is a lower bound. Among the leading models, additional spending yields no measurable gain in reward on this benchmark.

\providecommand{\rankone}[1]{{\color[HTML]{0A46D8}\normalsize\bfseries #1}}
\providecommand{\ranktwo}[1]{{\color[HTML]{4169E1}\small\bfseries #1}}
\providecommand{\rankthree}[1]{{\color[HTML]{8442C4}\small #1}}
\providecommand{\subcell}[1]{{\footnotesize\color[HTML]{6B6478}#1}}
\definecolor{rowtint}{HTML}{F1EDF7}
\definecolor{accentblue}{HTML}{0A46D8}

\begin{table}[!ht]
\centering
\begin{tblr}{
  colspec = {X[l,2.5,m] X[c,0.85] X[c,0.85] X[c,0.85] X[c,0.85] X[c,0.85] X[c,1.0] X[c,1.05]},
  width = \linewidth,
  row{1-2} = {font=\small},
  column{1} = {font=\small},
  row{3-Z} = {font=\footnotesize},
  hline{1,Z} = {0.9pt},
}
\SetCell[r=2]{l,m} Model & \SetCell{c} C0 & \SetCell{c} C1 & \SetCell{c} C2 & \SetCell{c} C3 & \SetCell{c} C4 & \SetCell{c} overall & \SetCell[r=2]{c,m} cost \\
 & \subcell{judged} & \subcell{Jaccard} & \subcell{Jaccard} & \subcell{ran $\geq$2} & \subcell{query F1} & \subcell{reward} & \\
\mbox{GLM-5.3-flash} & 0.88 & \rankone{0.73} & 0.87 & \ranktwo{0.98} & \ranktwo{0.62} & \rankone{0.74} & \subcell{\$2} \\
\mbox{Kimi-K3} & 0.86 & \ranktwo{0.72} & 0.85 & \ranktwo{0.98} & \rankthree{0.58} & \ranktwo{0.72} & \subcell{\$32} \\
\mbox{Grok-4.6} & 0.57 & 0.49 & 0.87 & \ranktwo{0.98} & \rankone{0.68} & \rankthree{0.71} & \subcell{\$28} \\
\mbox{GLM-5.3} & 0.90 & 0.70 & 0.86 & \rankthree{0.96} & 0.54 & 0.70 & \subcell{\$20} \\
\mbox{Claude-Opus-5} & \rankone{0.96} & \rankthree{0.71} & 0.85 & \rankone{1.00} & 0.50 & 0.69 & \subcell{\$50} \\
\mbox{Qwen3.8-27B} & \rankthree{0.93} & \rankone{0.73} & 0.88 & \ranktwo{0.98} & 0.50 & 0.68 & \subcell{\$10} \\
\mbox{GPT-5.6-sol} & \ranktwo{0.95} & 0.43 & 0.87 & 0.80 & 0.54 & 0.68 & \subcell{\$9} \\
\mbox{Mistral-Large-4} & \ranktwo{0.95} & \ranktwo{0.72} & 0.89 & \rankone{1.00} & 0.47 & 0.68 & \subcell{\$22} \\
\mbox{GLM-5.2} & 0.87 & 0.67 & \rankthree{0.90} & 0.92 & 0.48 & 0.66 & \subcell{\$13} \\
\mbox{Qwen3.8-2.4T} & 0.92 & 0.70 & 0.87 & \rankone{1.00} & 0.44 & 0.66 & \subcell{\$17} \\
\mbox{DeepSeek-v4-pro} & 0.86 & 0.66 & 0.89 & \ranktwo{0.98} & 0.45 & 0.65 & \subcell{\$24} \\
\mbox{GPT-5.6-terra} & 0.88 & 0.34 & 0.89 & 0.52 & 0.54 & 0.65 & \subcell{\$3} \\
\mbox{DeepSeek-v4.1-flash} & 0.88 & 0.67 & \ranktwo{0.91} & \rankone{1.00} & 0.43 & 0.64 & \subcell{\$5} \\
\mbox{Claude-Sonnet-5} & 0.88 & 0.64 & 0.87 & \rankthree{0.96} & 0.42 & 0.63 & \subcell{\$31} \\
\mbox{DeepSeek-v4-flash} & 0.88 & 0.58 & \rankone{0.93} & \rankthree{0.96} & 0.44 & 0.63 & \subcell{\$1} \\
\mbox{MiniMax-M3} & 0.87 & 0.63 & 0.87 & 0.90 & 0.36 & 0.58 & \subcell{\$5} \\
\mbox{Inkling} & 0.75 & 0.46 & 0.87 & 0.82 & 0.39 & 0.57 & \subcell{\$9} \\
\mbox{Nemotron-3-Ultra} & 0.73 & 0.48 & 0.88 & 0.78 & 0.36 & 0.56 & \subcell{\$17} \\
\mbox{GPT-5.6-luna} & 0.86 & 0.37 & \rankthree{0.90} & 0.56 & 0.35 & 0.55 & \subcell{28\textcent{}} \\
\mbox{Gemma-4-31B} & 0.73 & 0.23 & 0.87 & 0.54 & 0.38 & 0.52 & \subcell{70\textcent{}} \\
\mbox{Inkling-Small} & 0.70 & 0.30 & 0.79 & 0.50 & 0.36 & 0.50 & \subcell{\$2} \\
\mbox{Qwen3.6-35B-A3B} & 0.84 & 0.42 & 0.89 & 0.82 & 0.24 & 0.49 & \subcell{\$2} \\
\mbox{Claude-Haiku-4.5} & 0.49 & 0.42 & 0.89 & 0.54 & 0.33 & 0.48 & \subcell{\$7} \\
\mbox{Gemma-4-26B-A4B} & 0.57 & 0.26 & 0.84 & 0.48 & 0.35 & 0.47 & \subcell{\$1} \\
\mbox{GPT-OSS-120B} & 0.49 & 0.20 & 0.74 & 0.16 & 0.18 & 0.34 & \subcell{\$3} \\
\mbox{GPT-OSS-20B} & 0.53 & 0.14 & 0.61 & 0.16 & 0.11 & 0.27 & \subcell{51\textcent{}} \\
\end{tblr}
\par\vspace{6pt}
\caption{Results for all 26 models at the \emph{unconstrained budget} of 300 agent messages, ordered by overall reward. Columns C0 to C4 are the benchmark's own five checkpoint scores, each on its own 0--1 scale: C0 is the judged score for reading the threat-intelligence report, C1 and C2 are the Jaccard scores for the ATT\&CK techniques and the telemetry tables, C3 is the share of tasks in which at least two queries returned rows, and C4 is the F1 of the detection query. \emph{Overall} is the weighted reward of Eq.~\eqref{eq:reward}; it also counts the judged quality of the Sigma rule, which has no column here. No sample reaches the message cap at this budget. \emph{Cost} is the total cost of the run. In each score column the three best values are marked {\color[HTML]{0A46D8}first}, {\color[HTML]{4169E1}second} and {\color[HTML]{8442C4}third}, ranked on the precision shown so that equal printed values share a rank; the cost column, for which a lower value is better, is left unranked.}
\label{tab:cti_realm_bunlim}
\end{table}

The two figures have different leading models. We observe that at the unconstrained budget the highest reward is that of GLM-5.3-flash, 0.74 at \$2, and that at the published limit it is that of Kimi-K3, 0.69 at \$27. Both leaders are open-weight. At the published limit Kimi-K3 leads the best closed model, GPT-5.6-sol, by 0.010, with a 95\% interval of [\textminus{}0.042, 0.055] from the paired bootstrap described next. The lead at the unconstrained budget is also narrow: GLM-5.3-flash is 0.03 ahead of Kimi-K3, which ranks second. A paired bootstrap (10{,}000 replicates, resampling the 43 emulated intrusions the 50 tasks come from, so that tasks from one intrusion stay together) puts the gap at 0.025, with a 95\% interval of [0.002, 0.057]. This interval reflects which tasks were sampled, not the variation between runs of the same model. The gap also depends on which of GLM-5.3-flash's runs is reported. Figure~\ref{fig:cost_accuracy_b300} and Table~\ref{tab:cti_realm_bunlim} report its budget-150 run with the two truncated samples re-run at 300. Its run made entirely at budget 300 scores 0.703, which would place it third, 0.010 behind Grok-4.6 (interval [\textminus{}0.062, 0.027]) and 0.014 behind Kimi-K3 (interval [\textminus{}0.051, 0.015]), intervals that both include zero. Kimi-K3, Claude-Opus-5 and Grok-4.6 have one run each above the published limit. We therefore regard the leading models as sharing the top of the ranking. The cost gap is far larger: at the unconstrained budget Kimi-K3's run costs 19 times as much as GLM-5.3-flash's.

The message cap is why we report the unconstrained budget. A truncated sample never reaches \texttt{submit()} and scores zero on every checkpoint, including ones it had already passed, so at the published limit a model with a dozen truncated samples is measured on termination as much as on capability. DeepSeek-v4-flash is the clearest case: 12 of its 50 samples were truncated at 70 messages and none at 250, the budget from which its unconstrained result is carried, and its reward rises from 0.51 to 0.62, the largest gain of any model after GLM-5.3-flash's. A comparison at the published limit therefore measures termination and capability together, which is why we read the results at a budget where no reported sample is truncated.

\emph{Price and capability among the leading models.} The ten highest-scoring models span 0.66 to 0.74 in reward. Their runs cost \$2 to \$50. We observe that within this group the cost of a run varies by more than an order of magnitude: GLM-5.3-flash at \$2 and Claude-Opus-5 at \$50 differ by 0.05 in reward (0.743 against 0.690). Note that we do not claim price never tracks capability. Across all 26 models it does (Spearman correlation 0.64 between cost and reward); among the ten leading models it does not (0.14). The claim is narrower and more useful to a buyer: among the models good enough to consider for this task, the price of a model is not evidence about which one to choose.

\subsection{Output Token Usage}
\label{sec:tokens}

\begin{figure}[!ht]
\centering
\includegraphics[width=\linewidth]{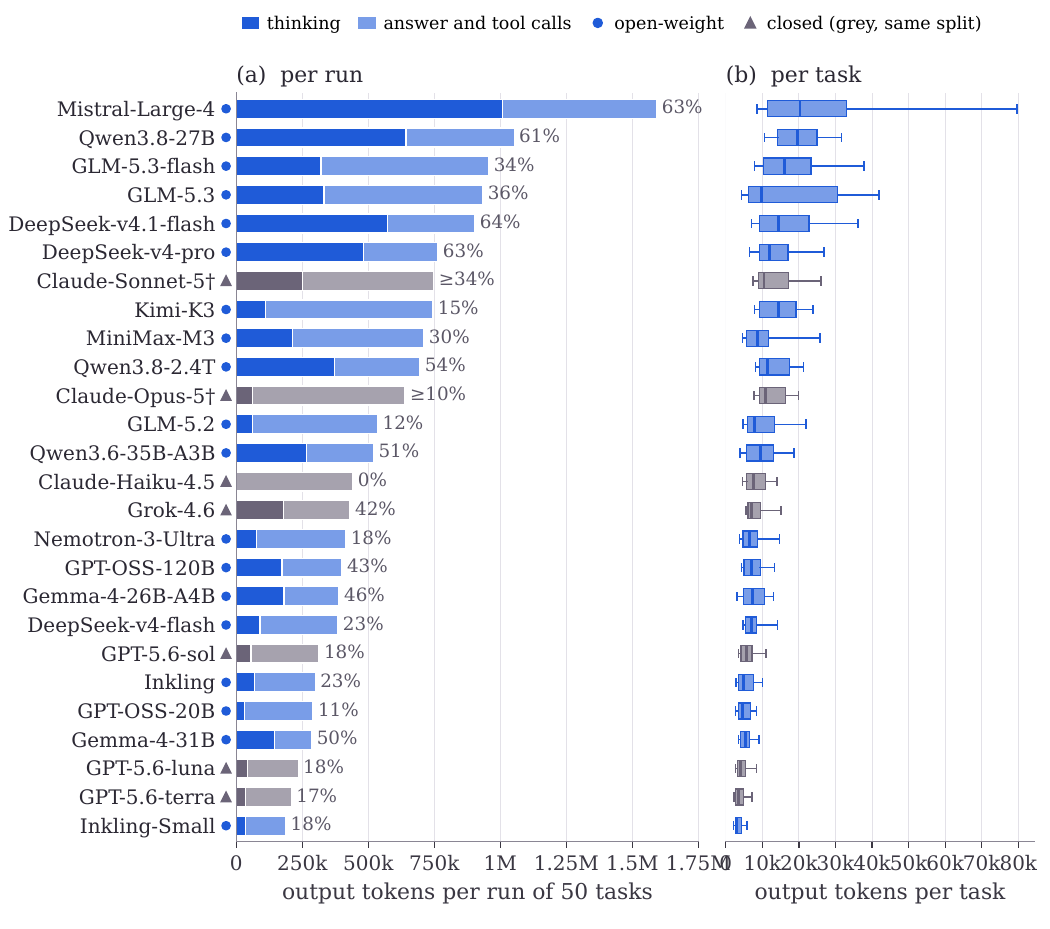}
\caption{\emph{Tokens generated per run and per task, at the unconstrained budget.} \emph{(a)} output tokens summed over the 50 tasks, split into thinking (dark) and the answer and tool calls (light); the number at the end of each bar is the thinking share, the fraction of output tokens spent on thinking. Blue marks open-weight models and grey closed ones, with the marker shapes of Figure~\ref{fig:cost_accuracy_b300}. \emph{(b)} the same output spread over the 50 tasks: the box spans the middle half of the tasks, the line is the median and the whiskers reach the 10th and 90th percentiles. The carried models are those drawn with hollow markers in Figure~\ref{fig:cost_accuracy_b300}. \textdagger{} marks a thinking share that is a lower bound. Scorer and judge-panel calls are excluded.}
\label{fig:tokens}
\end{figure}

Figure~\ref{fig:tokens} counts the output tokens each model generates at the unconstrained budget, thinking included: per run in panel (a) and per task in panel (b). A count is the reported output-token total, summed over every model call in the 50 trajectories behind the reward in Table~\ref{tab:cti_realm_bunlim}; calls made by the benchmark's in-run scorer and by the judge panel are excluded. We verify rather than assume that thinking is included in that count. On the 19 models that return their thinking as text, the visible thinking, answer and tool-call text comes to 3.6 to 4.2 characters per counted token, the usual rate for English text and code, and the reported thinking exceeds the reported output on at most 0.36\% of any run's output. Grok-4.6, GPT-5.6-sol, GPT-5.6-terra and GPT-5.6-luna return their thinking encrypted; for these models the reported output minus the reported thinking matches the visible text at 3.9 to 4.2 characters per token, so the reported thinking is exactly the hidden part. Claude-Opus-5 and Claude-Sonnet-5 return no thinking text, so their split cannot be checked and we report their thinking share as a lower bound.

We observe that a run of 50 tasks generates 187k (Inkling-Small) to 1.59M (Mistral-Large-4) output tokens, an 8.5-fold range, and that volume tracks reward only moderately (Spearman correlation 0.59 over the 26 models). The ten leading models reach their rewards at very different volumes: among them, output varies 5.1-fold, from 313k (GPT-5.6-sol) to 1.59M (Mistral-Large-4). Note that the relation does hold at the low end: the 6 runs that generate least have a median reward of 0.54, against 0.65 for the rest. Low volume is associated with a low reward, but high volume is not associated with a high one.

Thinking takes 11\% (GPT-OSS-20B) to 64\% (DeepSeek-v4.1-flash) of the output among the 23 models whose share is measured, with lower bounds of 10\% and 34\% for Claude-Opus-5 and Claude-Sonnet-5, and Claude-Haiku-4.5 does not think at all. The thinking share says little about the reward (Spearman correlation 0.07 over the 26 models). Among the leading models, GLM-5.2 spends 12\% of its output thinking and Mistral-Large-4 63\%, and the two differ by 0.02 in reward. The thinking share describes how a model works, not how well it performs on this benchmark.

Panel (b) shows that a run's output is not spread evenly over its tasks. For the median model the 90th-percentile task generates 3.2 times the output of the 10th-percentile task, and the five heaviest tasks, a tenth of the run, account for 22\% of its output. Per-task medians range from 3k (Inkling-Small) to 20k (Mistral-Large-4). GLM-5.3 has the widest spread, a ratio of 9.6 between those two percentiles, and MiniMax-M3 the most concentrated run, with 44\% of its output in its five heaviest tasks. A total per run therefore hides a long tail of expensive tasks.

Output is a small part of what a run processes. Each turn re-reads the growing transcript, so the input of a run comes to 10M to 62M tokens and output is 1.1\% to 3.2\% of all tokens processed; 4\% to 90\% of the input is read from cache. The cost of a run is therefore set by the price per token far more than by how much the model writes: GLM-5.3-flash generates 1.5 times as much output as Claude-Opus-5, and its run costs \$2 against \$50.

\section{Results by Checkpoint}
\label{sec:stages}

At the unconstrained budget, each checkpoint's table gives the benchmark's scores, diagnostics we compute with its definitions and, for C1 to C4, our metrics, which score the agent's submitted KQL query and Sigma rule. They complement the benchmark's measures of exploration, querying and the form of the rule for C2, C3 and the Sigma rule, and correct how it measures the same quantity for C1 and the C4 query. They are computed from the stored trajectories without re-running any agent, do not enter the reward, and average per-task values, so a model's F1 can lie below both its precision and its recall. For C1 to C4, a figure draws the key metrics of the stage, the benchmark's and ours, for the five models that rank highest on it; the table keeps every metric. Of these, C1, the Sigma rule, the C4 queries whose rows cannot be matched to events and the C2 tables that the key's reference queries leave open are graded as in Eq.~\eqref{eq:panel} by Gemma-4-31B with reasoning enabled, DeepSeek-v4.1-flash and GLM-5.3-flash, the reward's panel with DeepSeek-v4.1-flash in place of DeepSeek-v4-flash; each judge is also an evaluated model.

\subsection{C0: Threat-Intelligence Research}
\label{sec:stage_c0}

C0 asks the agent to find a relevant threat-intelligence report and explain why; the benchmark's judge prompt, applied by the reward's panel (Section~\ref{sec:setup}), grades the research's relevance to the detection objective. The answer key holds no reference for C0, and in 18 of the 50 tasks no report in the corpus is judged by both GLM-5.3-flash and DeepSeek-v4.1-flash to describe the target behaviour, so we add no metric of our own. Every model opens a report on at least 0.94 of its tasks, so access to the corpus does not explain the spread in the score; reports opened per task correlate moderately with it (Spearman correlation 0.57; Table~\ref{tab:stage_c0}).

\begin{table}[!tp]
\centering
\begin{tblr}{
  colspec = {X[l,1.55,m] X[c,1.0,m] X[c,1.0,m] X[c,1.0,m]},
  colsep = 3pt, width = \linewidth, rowsep = 1pt,
  row{1-2} = {font=\footnotesize}, row{3-Z} = {font=\footnotesize},
  hline{1,Z} = {0.9pt}, hline{2} = {2-Z}{0.4pt}, hline{3} = {0.4pt}, 
}
\SetCell[r=2]{l,m} Model & \SetCell[c=3]{c} Benchmark score and diagnostics & & \\
 & {score\\(judged)} & {opened any report} & {reports per task} \\
\mbox{GLM-5.3-flash} & 0.88 & 1.00 & 10.8 \\
\mbox{Kimi-K3} & 0.86 & 1.00 & 8.9 \\
\mbox{Grok-4.6} & 0.57 & 1.00 & 16.0 \\
\mbox{GLM-5.3} & 0.90 & 1.00 & 12.0 \\
\mbox{Claude-Opus-5} & \rankone{0.96} & 1.00 & 6.3 \\
\mbox{Qwen3.8-27B} & \rankthree{0.93} & 1.00 & 10.9 \\
\mbox{GPT-5.6-sol} & \ranktwo{0.95} & 1.00 & 6.9 \\
\mbox{Mistral-Large-4} & \ranktwo{0.95} & 1.00 & 13.0 \\
\mbox{GLM-5.2} & 0.87 & 1.00 & 7.5 \\
\mbox{Qwen3.8-2.4T} & 0.92 & 1.00 & 10.6 \\
\mbox{DeepSeek-v4-pro} & 0.86 & 1.00 & 11.4 \\
\mbox{GPT-5.6-terra} & 0.88 & 1.00 & 5.2 \\
\mbox{DeepSeek-v4.1-flash} & 0.88 & 1.00 & 18.3 \\
\mbox{Claude-Sonnet-5} & 0.88 & 1.00 & 11.1 \\
\mbox{DeepSeek-v4-flash} & 0.88 & 1.00 & 7.6 \\
\mbox{MiniMax-M3} & 0.87 & 0.98 & 5.5 \\
\mbox{Inkling} & 0.75 & 0.98 & 7.9 \\
\mbox{Nemotron-3-Ultra} & 0.73 & 1.00 & 6.9 \\
\mbox{GPT-5.6-luna} & 0.86 & 1.00 & 5.0 \\
\mbox{Gemma-4-31B} & 0.73 & 1.00 & 3.4 \\
\mbox{Inkling-Small} & 0.70 & 0.96 & 4.7 \\
\mbox{Qwen3.6-35B-A3B} & 0.84 & 1.00 & 7.4 \\
\mbox{Claude-Haiku-4.5} & 0.49 & 1.00 & 4.5 \\
\mbox{Gemma-4-26B-A4B} & 0.57 & 0.94 & 2.1 \\
\mbox{GPT-OSS-120B} & 0.49 & 0.94 & 3.0 \\
\mbox{GPT-OSS-20B} & 0.53 & 0.98 & 3.3 \\
\end{tblr}
\par\vspace{4pt}
\caption{C0, threat-intelligence research, at the unconstrained budget. Benchmark: the score of the agent's research, graded with the benchmark's judge prompt by the reward's panel (Section~\ref{sec:setup}), and two diagnostics we compute, the share of tasks in which the agent opened at least one report and the reports it opened per task. The answer key holds no reference for C0, so there is no metric of ours. Rows follow the benchmark's overall reward (Table~\ref{tab:cti_realm_bunlim}); colours mark the {\color[HTML]{0A46D8}first}, {\color[HTML]{4169E1}second} and {\color[HTML]{8442C4}third} best values of the score, equal values sharing a colour.}
\label{tab:stage_c0}
\end{table}

\begin{figure}[!tp]
\centering
\includegraphics[width=0.85\linewidth]{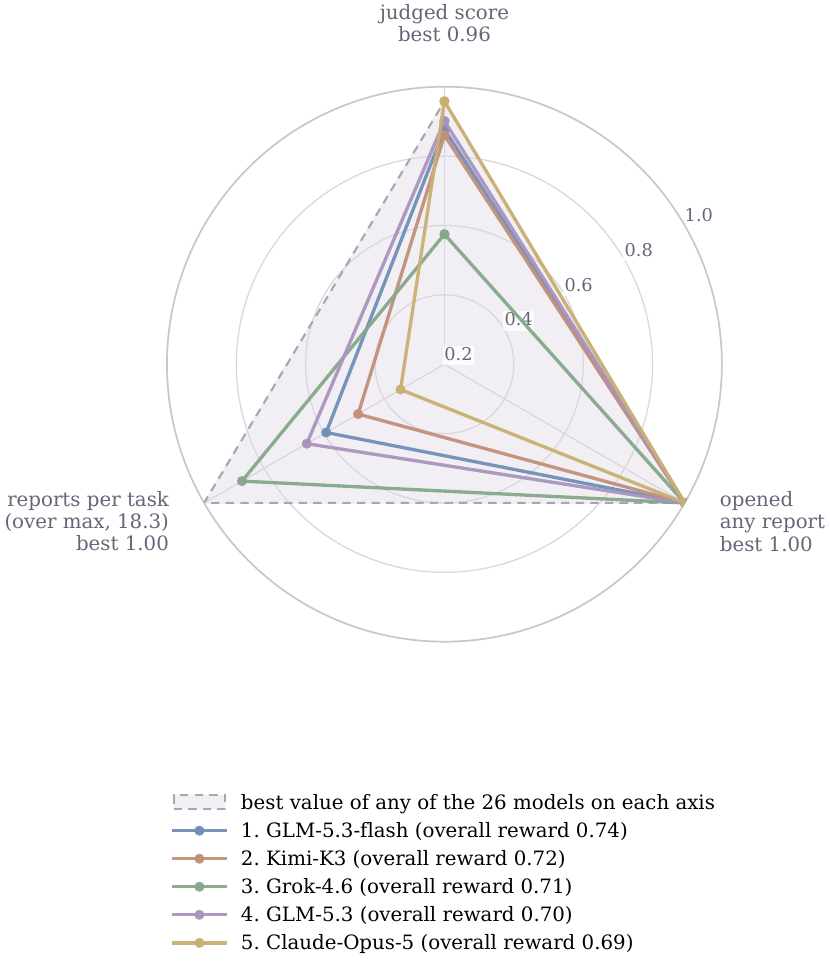}
\caption{The three columns of Table~\ref{tab:stage_c0} at the unconstrained budget, for the five models with the highest overall reward, in the colours of the figures for C1 to C4; reports per task are divided by the largest value any model reaches. The shaded hull joins the best value that any of the 26 models reaches on each axis, which is printed under the axis label.}
\label{fig:radar_c0}
\end{figure}

\FloatBarrier   %
\subsection{C1: ATT\&CK Technique Mapping}
\label{sec:stage_c1}

\begin{figure}[!tp]
\centering
\includegraphics[width=0.85\linewidth]{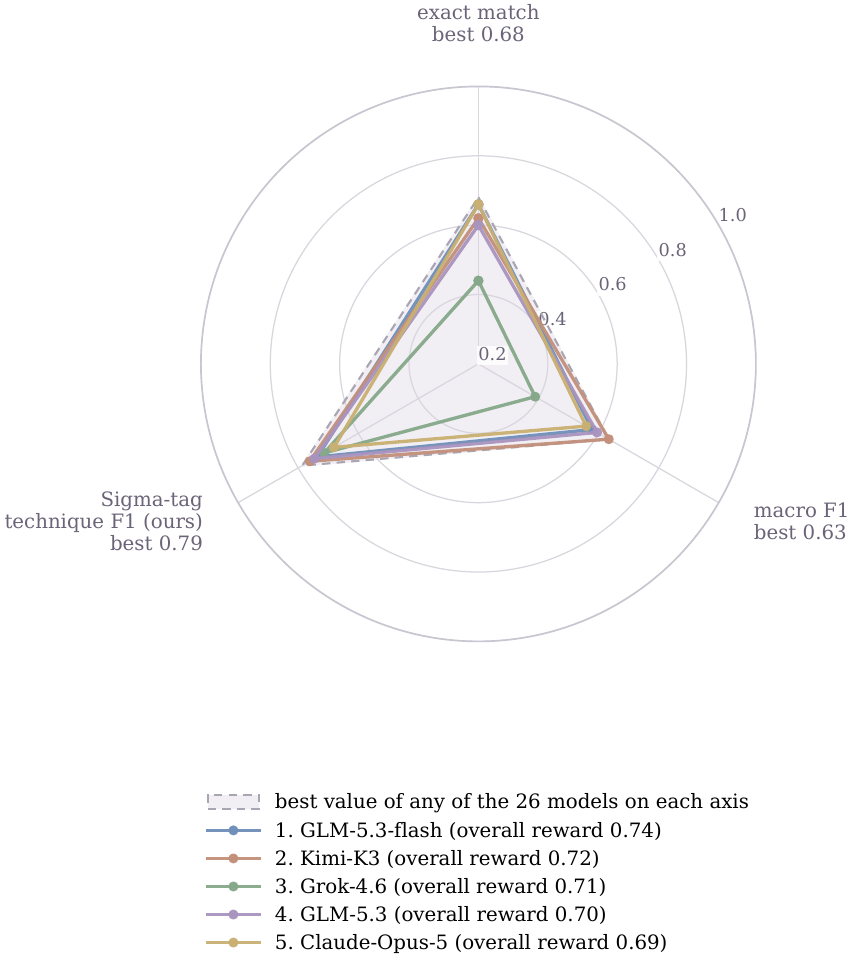}
\caption{Key metrics of C1 at the unconstrained budget, the benchmark's and ours (marked ``ours''), for the five models with the highest overall reward, the same five in the same colours in every checkpoint's figure. The shaded hull joins the best value that any of the 26 models reaches on each axis, which is printed under the axis label. Table~\ref{tab:stage_c1} gives every metric for every model.}
\label{fig:radar_c1}
\end{figure}

C1 concerns the mapping of the campaign to ATT\&CK techniques, whose identifiers the task never asks for. The benchmark nonetheless collects the key's identifiers anywhere in the agent's output, reasoning included, so its Jaccard index with the key equals the recall of the key: a wrong identifier costs nothing, and with exact matching a sub-technique of a key technique earns no credit. We score instead the mapping the agent commits to, the techniques tagged in its submitted Sigma rule, as Sigma rules conventionally carry ATT\&CK tags; an untagged rule, about one in eight, scores zero. A tag counts in proportion to the judge trials that match it to a key technique, whatever its identifier; one the judges accept but the key omits is neither credited nor penalised, as the key gives a single technique for 38 of the 50 tasks. A judge repeats its verdict in all three trials for 89 to 93\% of tags, and the judges' majority verdicts agree on 89\%. The three GPT-5.6 models score 0.23 to 0.36 higher on our F1 than on the benchmark's score (Table~\ref{tab:stage_c1} and Figure~\ref{fig:radar_c1}).

\begin{table}[!tp]
\centering
\begin{tblr}{
  colspec = {X[l,2.15,m] X[c,1.0,m] X[c,1.0,m] X[c,1.0,m] X[c,1.0,m] X[c,1.0,m] X[c,1.0,m]},
  colsep = 3pt, width = \linewidth, rowsep = 1pt,
  row{1-2} = {font=\footnotesize}, row{3-Z} = {font=\footnotesize},
  hline{1,Z} = {0.9pt}, hline{2} = {2-Z}{0.4pt}, hline{3} = {0.4pt}, vline{5} = {2-Z}{0.4pt},
}
\SetCell[r=2]{l,m} Model & \SetCell[c=3]{c} Benchmark score and diagnostics & & & \SetCell[c=3]{c} Sigma-rule tags, judged (ours) & & \\
 & {Jaccard\\(key recall)} & {exact match} & {macro F1} & {precision} & {recall} & {F1\\(Sigma rule's techniques)} \\
\mbox{GLM-5.3-flash} & \rankone{0.73} & \ranktwo{0.66} & 0.58 & 0.72 & 0.83 & 0.74 \\
\mbox{Kimi-K3} & \ranktwo{0.72} & 0.62 & \rankone{0.63} & 0.79 & 0.79 & \rankthree{0.76} \\
\mbox{Grok-4.6} & 0.49 & 0.44 & 0.39 & 0.71 & 0.78 & 0.71 \\
\mbox{GLM-5.3} & 0.70 & 0.60 & \rankthree{0.59} & 0.75 & 0.85 & 0.75 \\
\mbox{Claude-Opus-5} & \rankthree{0.71} & \ranktwo{0.66} & 0.56 & 0.62 & 0.85 & 0.68 \\
\mbox{Qwen3.8-27B} & \rankone{0.73} & \rankone{0.68} & 0.47 & 0.63 & 0.69 & 0.63 \\
\mbox{GPT-5.6-sol} & 0.43 & 0.38 & 0.41 & 0.83 & 0.73 & 0.74 \\
\mbox{Mistral-Large-4} & \ranktwo{0.72} & \ranktwo{0.66} & 0.58 & 0.79 & 0.82 & \ranktwo{0.78} \\
\mbox{GLM-5.2} & 0.67 & 0.60 & 0.53 & 0.62 & 0.72 & 0.64 \\
\mbox{Qwen3.8-2.4T} & 0.70 & \rankthree{0.64} & 0.54 & 0.81 & 0.81 & \rankone{0.79} \\
\mbox{DeepSeek-v4-pro} & 0.66 & 0.56 & 0.58 & 0.78 & 0.70 & 0.71 \\
\mbox{GPT-5.6-terra} & 0.34 & 0.28 & 0.33 & 0.80 & 0.67 & 0.70 \\
\mbox{DeepSeek-v4.1-flash} & 0.67 & 0.60 & \ranktwo{0.60} & 0.72 & 0.82 & 0.73 \\
\mbox{Claude-Sonnet-5} & 0.64 & 0.58 & 0.54 & 0.77 & 0.81 & \rankthree{0.76} \\
\mbox{DeepSeek-v4-flash} & 0.58 & 0.52 & 0.47 & 0.70 & 0.76 & 0.69 \\
\mbox{MiniMax-M3} & 0.63 & 0.58 & 0.45 & 0.60 & 0.69 & 0.60 \\
\mbox{Inkling} & 0.46 & 0.40 & 0.33 & 0.62 & 0.62 & 0.59 \\
\mbox{Nemotron-3-Ultra} & 0.48 & 0.44 & 0.39 & 0.61 & 0.61 & 0.59 \\
\mbox{GPT-5.6-luna} & 0.37 & 0.34 & 0.31 & 0.63 & 0.61 & 0.60 \\
\mbox{Gemma-4-31B} & 0.23 & 0.22 & 0.17 & 0.34 & 0.30 & 0.30 \\
\mbox{Inkling-Small} & 0.30 & 0.26 & 0.22 & 0.52 & 0.44 & 0.44 \\
\mbox{Qwen3.6-35B-A3B} & 0.42 & 0.36 & 0.41 & 0.43 & 0.43 & 0.40 \\
\mbox{Claude-Haiku-4.5} & 0.42 & 0.40 & 0.29 & 0.50 & 0.55 & 0.52 \\
\mbox{Gemma-4-26B-A4B} & 0.26 & 0.24 & 0.19 & 0.28 & 0.28 & 0.26 \\
\mbox{GPT-OSS-120B} & 0.20 & 0.16 & 0.17 & 0.15 & 0.14 & 0.14 \\
\mbox{GPT-OSS-20B} & 0.14 & 0.14 & 0.12 & 0.06 & 0.06 & 0.06 \\
\end{tblr}
\par\vspace{4pt}
\caption{C1, ATT\&CK technique mapping, at the unconstrained budget. Benchmark: the benchmark's Jaccard index between the key and the key's technique identifiers found anywhere in the agent's output, which therefore equals the recall of the key, and two diagnostics we compute, the share of tasks matched exactly and the macro F1 (F1 per technique class, averaged with equal weight). Sigma-rule technique tags (ours): F1 scores the technique mapping the agent commits to, the ATT\&CK techniques tagged in its submitted Sigma rule, against the key. In each task, precision is the share of the rule's tags that the judges match to a key technique, whatever the identifier; recall is the share of the key's techniques that some tag matches; F1 is their harmonic mean. A tag counts in proportion to the trials, three LLM judges in three trials each, that match it; a tag the judges accept but the key omits counts in neither; an untagged rule scores zero. Per-task values are averaged over the 50 tasks, so F1 can lie below both precision and recall. Rows follow the benchmark's overall reward (Table~\ref{tab:cti_realm_bunlim}); colours mark the {\color[HTML]{0A46D8}first}, {\color[HTML]{4169E1}second} and {\color[HTML]{8442C4}third} best values of the Jaccard, the exact match, the macro F1 and our F1, equal values sharing a colour.}
\label{tab:stage_c1}
\end{table}

\FloatBarrier   %
\subsection{C2: Telemetry Table Discovery}
\label{sec:stage_c2}

\begin{figure}[!tp]
\centering
\includegraphics[width=0.85\linewidth]{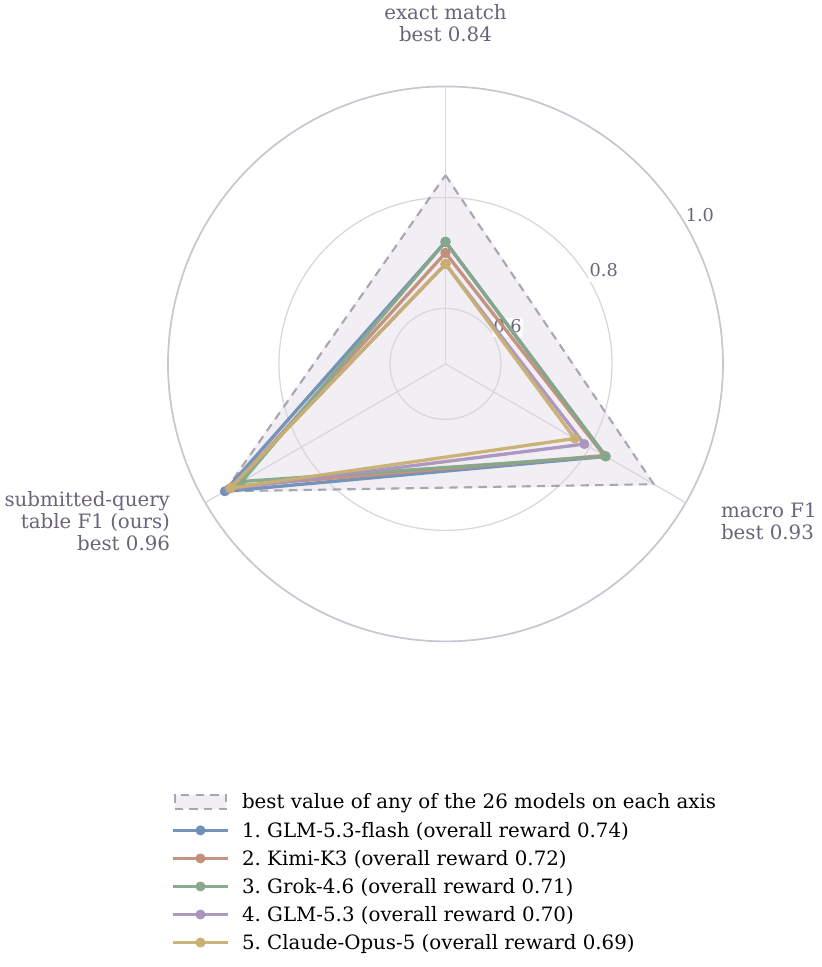}
\caption{Key metrics of C2 at the unconstrained budget, the benchmark's and ours (marked ``ours''), for the five models with the highest overall reward, the same five in the same colours in every checkpoint's figure. The shaded hull joins the best value that any of the 26 models reaches on each axis, which is printed under the axis label. Table~\ref{tab:stage_c2} gives every metric for every model.}
\label{fig:radar_c2}
\end{figure}

C2 asks the agent to find the telemetry tables that hold the evidence. The benchmark's Jaccard index is again the recall of the key, here over the tables the agent passed to its two exploration tools, a set it never commits to, so inspecting all 12 tables scores 1 and a table queried without being inspected earns nothing. We score the tables the submitted query reads by precision, recall and F1 against the tables that record the attack in the store. A key table is one of them when its reference query returns events or, since a reference query can miss events that are there, when the judges, shown the rows of that table that meet the key's conditions or carry its identifiers, find the attack in them. Of the 16 key tables whose reference queries return nothing, the judges find the attack in 9; in the two tasks on restricted administrative units, for instance, the attack's AuditLogs events leave empty the actor field on which the reference query filters. A table that holds copies of a key table's events, with the same event identifiers, stands in for it, as AKSAuditAdmin does for AKSAudit in 10 of the 16 AKS tasks in which a submission reads it; a table outside the key counts in proportion to the judge trials that find the attack in it and against precision otherwise, and is never required. The judges' majority verdicts agree on 84\% of the 19 tables they grade. Our F1 is at least 0.82 for every model except GPT-OSS-20B, at 0.48 (Table~\ref{tab:stage_c2} and Figure~\ref{fig:radar_c2}).

\begin{table}[!tp]
\centering
\begin{tblr}{
  colspec = {X[l,2.15,m] X[c,1.0,m] X[c,1.0,m] X[c,1.0,m] X[c,1.0,m] X[c,1.0,m] X[c,1.0,m]},
  colsep = 3pt, width = \linewidth, rowsep = 1pt,
  row{1-2} = {font=\footnotesize}, row{3-Z} = {font=\footnotesize},
  hline{1,Z} = {0.9pt}, hline{2} = {2-Z}{0.4pt}, hline{3} = {0.4pt}, vline{5} = {2-Z}{0.4pt},
}
\SetCell[r=2]{l,m} Model & \SetCell[c=3]{c} Benchmark score and diagnostics & & & \SetCell[c=3]{c} Tables queried, judged (ours) & & \\
 & {Jaccard\\(key recall)} & {exact match} & {macro F1} & {precision} & {recall} & {F1\\(submitted query's tables)} \\
\mbox{GLM-5.3-flash} & 0.87 & 0.72 & 0.83 & 1.00 & 0.94 & \rankone{0.96} \\
\mbox{Kimi-K3} & 0.85 & 0.70 & 0.83 & 1.00 & 0.91 & \rankthree{0.94} \\
\mbox{Grok-4.6} & 0.87 & 0.72 & 0.83 & 0.98 & 0.90 & 0.92 \\
\mbox{GLM-5.3} & 0.86 & 0.68 & 0.79 & 0.99 & 0.93 & \ranktwo{0.95} \\
\mbox{Claude-Opus-5} & 0.85 & 0.68 & 0.77 & 1.00 & 0.92 & \ranktwo{0.95} \\
\mbox{Qwen3.8-27B} & 0.88 & 0.74 & 0.85 & 0.98 & 0.89 & 0.92 \\
\mbox{GPT-5.6-sol} & 0.87 & 0.72 & 0.81 & 0.96 & 0.92 & 0.92 \\
\mbox{Mistral-Large-4} & 0.89 & \rankthree{0.78} & 0.88 & 0.96 & 0.94 & \rankthree{0.94} \\
\mbox{GLM-5.2} & \rankthree{0.90} & \rankthree{0.78} & \rankthree{0.89} & 0.98 & 0.92 & \rankthree{0.94} \\
\mbox{Qwen3.8-2.4T} & 0.87 & 0.72 & 0.81 & 0.98 & 0.89 & 0.92 \\
\mbox{DeepSeek-v4-pro} & 0.89 & 0.76 & 0.87 & 0.99 & 0.92 & \rankthree{0.94} \\
\mbox{GPT-5.6-terra} & 0.89 & 0.76 & 0.86 & 0.99 & 0.92 & \rankthree{0.94} \\
\mbox{DeepSeek-v4.1-flash} & \ranktwo{0.91} & \ranktwo{0.80} & \ranktwo{0.91} & 0.99 & 0.92 & \rankthree{0.94} \\
\mbox{Claude-Sonnet-5} & 0.87 & 0.74 & 0.83 & 0.99 & 0.90 & 0.93 \\
\mbox{DeepSeek-v4-flash} & \rankone{0.93} & \rankone{0.84} & \rankone{0.93} & 1.00 & 0.91 & \rankthree{0.94} \\
\mbox{MiniMax-M3} & 0.87 & 0.74 & 0.86 & 0.98 & 0.89 & 0.92 \\
\mbox{Inkling} & 0.87 & 0.74 & 0.78 & 0.96 & 0.87 & 0.90 \\
\mbox{Nemotron-3-Ultra} & 0.88 & 0.72 & 0.84 & 0.97 & 0.89 & 0.92 \\
\mbox{GPT-5.6-luna} & \rankthree{0.90} & \rankthree{0.78} & \rankthree{0.89} & 0.93 & 0.89 & 0.88 \\
\mbox{Gemma-4-31B} & 0.87 & 0.70 & 0.73 & 1.00 & 0.91 & \rankthree{0.94} \\
\mbox{Inkling-Small} & 0.79 & 0.64 & 0.52 & 0.94 & 0.85 & 0.88 \\
\mbox{Qwen3.6-35B-A3B} & 0.89 & 0.74 & 0.86 & 0.93 & 0.86 & 0.88 \\
\mbox{Claude-Haiku-4.5} & 0.89 & 0.74 & 0.85 & 1.00 & 0.88 & 0.92 \\
\mbox{Gemma-4-26B-A4B} & 0.84 & 0.70 & 0.68 & 0.97 & 0.88 & 0.91 \\
\mbox{GPT-OSS-120B} & 0.74 & 0.62 & 0.65 & 0.87 & 0.79 & 0.82 \\
\mbox{GPT-OSS-20B} & 0.61 & 0.50 & 0.39 & 0.54 & 0.45 & 0.48 \\
\end{tblr}
\par\vspace{4pt}
\caption{C2, telemetry table discovery, at the unconstrained budget. Benchmark: the benchmark's Jaccard index between the key and the key tables the agent passed to its two exploration tools, which therefore equals the recall of the key, and two diagnostics we compute, the share of tasks matched exactly and the macro F1 (F1 per table, averaged with equal weight). Tables read by the submitted query (ours): precision, recall and F1 of the tables the submitted query reads, against the tables that record the attack in the store: key tables whose reference queries return events or in which three LLM judges, in three trials each, find the attack, with copies of their events standing in for them, and tables outside the key in proportion to the judge trials that find the attack in them; per-task values averaged over the 50 tasks, so F1 can lie below both precision and recall. Rows follow the benchmark's overall reward (Table~\ref{tab:cti_realm_bunlim}); colours mark the {\color[HTML]{0A46D8}first}, {\color[HTML]{4169E1}second} and {\color[HTML]{8442C4}third} best values of the Jaccard, the exact match, the macro F1 and our F1, equal values sharing a colour.}
\label{tab:stage_c2}
\end{table}

\FloatBarrier   %
\subsection{C3: Query Execution}
\label{sec:stage_c3}

\begin{figure}[!tp]
\centering
\includegraphics[width=0.85\linewidth]{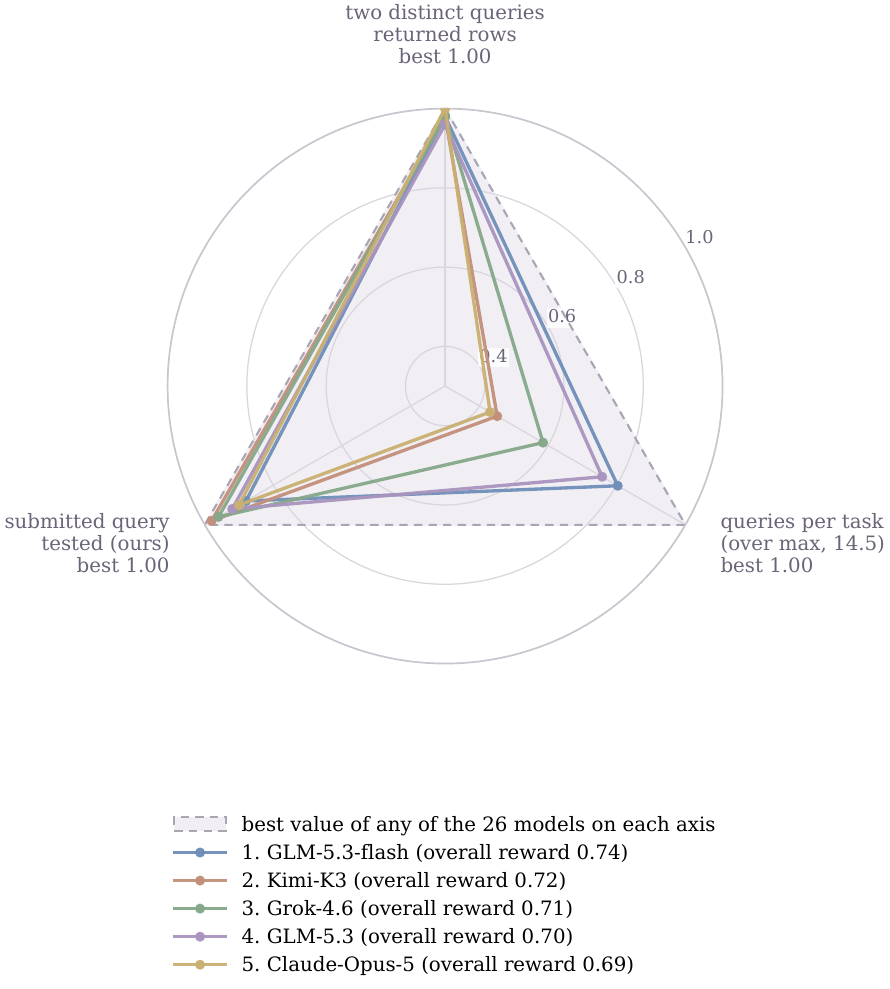}
\caption{Key metrics of C3 at the unconstrained budget, the benchmark's and ours (marked ``ours''), for the five models with the highest overall reward, the same five in the same colours in every checkpoint's figure. The shaded hull joins the best value that any of the 26 models reaches on each axis, which is printed under the axis label. Table~\ref{tab:stage_c3} gives every metric for every model. Queries per task are divided by the largest value any model reaches.}
\label{fig:radar_c3}
\end{figure}

C3 asks the agent to execute queries against the telemetry store. The benchmark passes a task when at least two distinct queries returned rows, which shows that the agent queried the store, not that it tested its submission. We report the share of tasks in which it did: an executed query with the submission's selection logic, ignoring formatting and operators that only shape the output, returned rows. A near miss, with a text similarity of at least 0.85 to the closest executed query that returned rows, also counts if both, run without row limits, return the same rows on at least two shared columns or an event identifier; 57 of the 81 near misses do. GPT-5.6-terra and GPT-5.6-luna pass the benchmark's check on 0.52 and 0.56 of their tasks but tested their submission on 1.00 and 0.98, while Nemotron-3-Ultra moves the other way, from 0.78 on the benchmark's check to 0.48 (Table~\ref{tab:stage_c3} and Figure~\ref{fig:radar_c3}).

\begin{table}[!tp]
\centering
\begin{tblr}{
  colspec = {X[l,1.55,m] X[c,1.0,m] X[c,1.0,m] X[c,1.0,m] X[c,1.0,m]},
  colsep = 3pt, width = \linewidth, rowsep = 1pt,
  row{1-2} = {font=\footnotesize}, row{3-Z} = {font=\footnotesize},
  hline{1,Z} = {0.9pt}, hline{2} = {2-Z}{0.4pt}, hline{3} = {0.4pt}, vline{5} = {2-Z}{0.4pt},
}
\SetCell[r=2]{l,m} Model & \SetCell[c=3]{c} Benchmark score and diagnostics & & & \SetCell[c=1]{c} Submitted query (ours) \\
 & {two distinct queries returned rows} & {queries per task} & {no query returned rows} & {submitted query tested} \\
\mbox{GLM-5.3-flash} & \ranktwo{0.98} & 11.7 & 0.00 & 0.88 \\
\mbox{Kimi-K3} & \ranktwo{0.98} & 6.6 & 0.00 & \ranktwo{0.98} \\
\mbox{Grok-4.6} & \ranktwo{0.98} & 8.5 & 0.00 & \rankthree{0.96} \\
\mbox{GLM-5.3} & \rankthree{0.96} & 11.0 & 0.00 & 0.92 \\
\mbox{Claude-Opus-5} & \rankone{1.00} & 6.3 & 0.00 & 0.90 \\
\mbox{Qwen3.8-27B} & \ranktwo{0.98} & 12.3 & 0.00 & \ranktwo{0.98} \\
\mbox{GPT-5.6-sol} & 0.80 & 4.7 & 0.00 & \rankone{1.00} \\
\mbox{Mistral-Large-4} & \rankone{1.00} & 14.5 & 0.00 & \rankone{1.00} \\
\mbox{GLM-5.2} & 0.92 & 9.0 & 0.00 & 0.90 \\
\mbox{Qwen3.8-2.4T} & \rankone{1.00} & 7.0 & 0.00 & \ranktwo{0.98} \\
\mbox{DeepSeek-v4-pro} & \ranktwo{0.98} & 8.9 & 0.00 & \rankone{1.00} \\
\mbox{GPT-5.6-terra} & 0.52 & 4.0 & 0.00 & \rankone{1.00} \\
\mbox{DeepSeek-v4.1-flash} & \rankone{1.00} & 12.0 & 0.00 & \rankthree{0.96} \\
\mbox{Claude-Sonnet-5} & \rankthree{0.96} & 8.9 & 0.00 & \rankthree{0.96} \\
\mbox{DeepSeek-v4-flash} & \rankthree{0.96} & 9.7 & 0.04 & 0.70 \\
\mbox{MiniMax-M3} & 0.90 & 11.2 & 0.02 & 0.78 \\
\mbox{Inkling} & 0.82 & 5.3 & 0.02 & 0.74 \\
\mbox{Nemotron-3-Ultra} & 0.78 & 6.9 & 0.02 & 0.48 \\
\mbox{GPT-5.6-luna} & 0.56 & 4.1 & 0.00 & \ranktwo{0.98} \\
\mbox{Gemma-4-31B} & 0.54 & 3.6 & 0.04 & 0.84 \\
\mbox{Inkling-Small} & 0.50 & 2.8 & 0.06 & 0.70 \\
\mbox{Qwen3.6-35B-A3B} & 0.82 & 7.3 & 0.04 & 0.74 \\
\mbox{Claude-Haiku-4.5} & 0.54 & 3.2 & 0.14 & 0.54 \\
\mbox{Gemma-4-26B-A4B} & 0.48 & 4.1 & 0.06 & 0.46 \\
\mbox{GPT-OSS-120B} & 0.16 & 2.5 & 0.10 & 0.62 \\
\mbox{GPT-OSS-20B} & 0.16 & 2.7 & 0.36 & 0.40 \\
\end{tblr}
\par\vspace{4pt}
\caption{C3, query execution, at the unconstrained budget. Benchmark: the share of tasks in which at least two distinct queries returned rows, and two diagnostics we compute, the queries executed per task and the share of tasks in which no query returned rows. Submitted query tested (ours): the share of the 50 tasks in which a query the agent executed, with the selection logic of its submitted KQL query, returned rows, a near miss counting when it returns the same rows. Rows follow the benchmark's overall reward (Table~\ref{tab:cti_realm_bunlim}); colours mark the {\color[HTML]{0A46D8}first}, {\color[HTML]{4169E1}second} and {\color[HTML]{8442C4}third} best values of the benchmark's score and of ours, equal values sharing a colour.}
\label{tab:stage_c3}
\end{table}

\FloatBarrier   %
\subsection{C4: Detection}
\label{sec:stage_c4}

C4 asks for the deliverable, a KQL query and a Sigma rule. The benchmark never runs the query: it computes an F1 over the key's field patterns on the rows the agent pasted into its answer, so the score depends on the output format, one fully matching row scores 1, and recall of the malicious events is never measured. We execute the submitted and reference queries in full in the telemetry store and report the precision, recall and F1 of the submission's events against theirs, on the 36 tasks in which every reference query returns events. Submissions returning counts or other aggregates, about one in ten, cannot be matched to events; the judges grade them against the key as equivalent (1), broader or narrower (0.5), or a miss or invalid (0), and the grade serves as their precision, recall and F1. Precision falls with events the reference queries do not return, recall with those a query misses; their medians are 0.64 and 0.74.

\begin{figure}[!tp]
\centering
\includegraphics[width=0.85\linewidth]{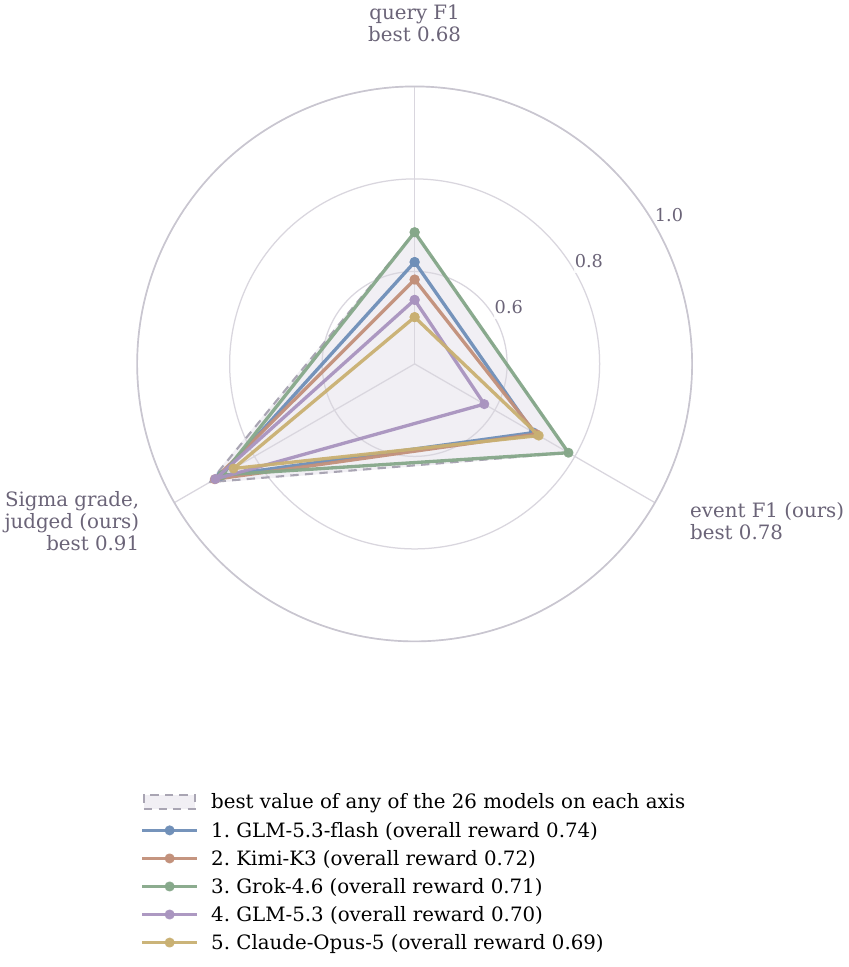}
\caption{Key metrics of C4 at the unconstrained budget, the benchmark's and ours (marked ``ours''), for the five models with the highest overall reward, the same five in the same colours in every checkpoint's figure. The shaded hull joins the best value that any of the 26 models reaches on each axis, which is printed under the axis label. Table~\ref{tab:stage_c4} gives every metric for every model.}
\label{fig:radar_c4}
\end{figure}

For the rule, the benchmark's judge prompt, applied by the reward's panel, rates syntax and specificity by a generic rubric that never consults the answer key. Our judges grade it against the key's reference queries and field patterns, on the scale used above for aggregate queries, without requiring the test environment's marker strings, account names or IP addresses. A judge repeats its grade in all three trials for 73 to 87\% of rules, and the judges' majority grades agree on 72\% (Table~\ref{tab:stage_c4} and Figure~\ref{fig:radar_c4}).

\begin{table}[!tp]
\centering
\begin{tblr}{
  colspec = {X[l,2.15,m] X[c,1.0,m] X[c,1.0,m] X[c,1.0,m] X[c,1.0,m] X[c,1.0,m] X[c,1.0,m] X[c,1.0,m]},
  colsep = 3pt, width = \linewidth, rowsep = 1pt,
  row{1-2} = {font=\footnotesize}, row{3-Z} = {font=\footnotesize},
  hline{1,Z} = {0.9pt}, hline{2} = {2-Z}{0.4pt}, hline{3} = {0.4pt}, vline{5} = {2-Z}{0.4pt},
}
\SetCell[r=2]{l,m} Model & \SetCell[c=3]{c} Benchmark scores and diagnostic & & & \SetCell[c=4]{c} Events and Sigma grade (ours) & & & \\
 & {query F1} & {zero query F1} & {Sigma\\(judged)} & {event precision} & {event recall} & {event F1} & {Sigma grade\\(judged)} \\
\mbox{GLM-5.3-flash} & \ranktwo{0.62} & 0.18 & \rankthree{0.88} & 0.68 & 0.85 & 0.70 & 0.88 \\
\mbox{Kimi-K3} & \rankthree{0.58} & 0.24 & 0.87 & 0.72 & 0.80 & 0.70 & \ranktwo{0.90} \\
\mbox{Grok-4.6} & \rankone{0.68} & 0.12 & 0.84 & 0.78 & 0.86 & \rankone{0.78} & 0.88 \\
\mbox{GLM-5.3} & 0.54 & 0.26 & 0.87 & 0.57 & 0.70 & 0.57 & \ranktwo{0.90} \\
\mbox{Claude-Opus-5} & 0.50 & 0.28 & \rankthree{0.88} & 0.72 & 0.80 & 0.71 & 0.85 \\
\mbox{Qwen3.8-27B} & 0.50 & 0.32 & 0.86 & 0.65 & 0.77 & 0.66 & 0.85 \\
\mbox{GPT-5.6-sol} & 0.54 & 0.30 & \rankone{0.90} & 0.78 & 0.78 & \rankthree{0.74} & \rankone{0.91} \\
\mbox{Mistral-Large-4} & 0.47 & 0.40 & 0.87 & 0.66 & 0.83 & 0.68 & 0.88 \\
\mbox{GLM-5.2} & 0.48 & 0.34 & 0.82 & 0.65 & 0.74 & 0.64 & 0.85 \\
\mbox{Qwen3.8-2.4T} & 0.44 & 0.32 & \ranktwo{0.89} & 0.64 & 0.73 & 0.63 & 0.88 \\
\mbox{DeepSeek-v4-pro} & 0.45 & 0.24 & 0.86 & 0.75 & 0.86 & \ranktwo{0.75} & 0.88 \\
\mbox{GPT-5.6-terra} & 0.54 & 0.32 & 0.87 & 0.74 & 0.79 & 0.70 & \rankthree{0.89} \\
\mbox{DeepSeek-v4.1-flash} & 0.43 & 0.28 & 0.86 & 0.64 & 0.78 & 0.65 & \ranktwo{0.90} \\
\mbox{Claude-Sonnet-5} & 0.42 & 0.40 & 0.85 & 0.72 & 0.77 & 0.71 & 0.86 \\
\mbox{DeepSeek-v4-flash} & 0.44 & 0.38 & 0.76 & 0.51 & 0.57 & 0.49 & 0.76 \\
\mbox{MiniMax-M3} & 0.36 & 0.52 & 0.77 & 0.59 & 0.65 & 0.56 & 0.74 \\
\mbox{Inkling} & 0.39 & 0.24 & 0.76 & 0.49 & 0.64 & 0.49 & 0.71 \\
\mbox{Nemotron-3-Ultra} & 0.36 & 0.50 & 0.81 & 0.52 & 0.58 & 0.51 & 0.75 \\
\mbox{GPT-5.6-luna} & 0.35 & 0.46 & 0.83 & 0.77 & 0.83 & \ranktwo{0.75} & 0.88 \\
\mbox{Gemma-4-31B} & 0.38 & 0.34 & 0.73 & 0.59 & 0.72 & 0.57 & 0.80 \\
\mbox{Inkling-Small} & 0.36 & 0.40 & 0.71 & 0.59 & 0.75 & 0.57 & 0.66 \\
\mbox{Qwen3.6-35B-A3B} & 0.24 & 0.54 & 0.67 & 0.43 & 0.60 & 0.44 & 0.68 \\
\mbox{Claude-Haiku-4.5} & 0.33 & 0.38 & 0.69 & 0.49 & 0.51 & 0.48 & 0.62 \\
\mbox{Gemma-4-26B-A4B} & 0.35 & 0.36 & 0.67 & 0.55 & 0.71 & 0.55 & 0.67 \\
\mbox{GPT-OSS-120B} & 0.18 & 0.60 & 0.64 & 0.41 & 0.45 & 0.40 & 0.55 \\
\mbox{GPT-OSS-20B} & 0.11 & 0.74 & 0.43 & 0.22 & 0.25 & 0.20 & 0.32 \\
\end{tblr}
\par\vspace{4pt}
\caption{C4, detection, at the unconstrained budget. Benchmark: the F1 over the key's field patterns of the rows the agent pasted into its answer, without running the query; a diagnostic we compute, the share of tasks whose query F1 is zero; and the quality of the Sigma rule, graded with the benchmark's judge prompt by the reward's panel (Section~\ref{sec:setup}). Executed submission and Sigma grade (ours): precision, recall and F1 of the events the submitted query returns when executed in full, against those of the key's reference queries, with submissions returning aggregates, about one in ten, graded by three LLM judges; per-task values averaged over the 36 tasks in which every reference query returns events, so F1 can lie below both precision and recall; and the grade of the Sigma rule against the key by the same judges, averaged over the 50 tasks. Rows follow the benchmark's overall reward (Table~\ref{tab:cti_realm_bunlim}); colours mark the {\color[HTML]{0A46D8}first}, {\color[HTML]{4169E1}second} and {\color[HTML]{8442C4}third} best values of the query F1, both Sigma columns and the event F1, equal values sharing a colour.}
\label{tab:stage_c4}
\end{table}

\FloatBarrier   %
\subsection{Effect on the Comparison}
\label{sec:stage_effect}

Comparing medians, open-weight against closed, our metrics reverse the open-weight lead on the benchmark's C1 (0.63 against 0.43) and C3 (0.92 against 0.80), to 0.63 against 0.70 and 0.84 against 0.96, and widen the closed models' existing lead on both parts of C4, from 0.43 against 0.50 to 0.57 against 0.71 on the query and from 0.81 against 0.85 to 0.80 against 0.88 on the rule; on C2 the families are level (0.92 against 0.92). On each metric the best open-weight and the best closed model differ by at most 0.04, which should not be read as an ordering: open-weight models lead C1 (Qwen3.8-2.4T at 0.79 against 0.76) and C2 (GLM-5.3-flash at 0.96 against 0.95), Mistral-Large-4 and DeepSeek-v4-pro share 1.00 on C3 with GPT-5.6-sol and GPT-5.6-terra, and closed models lead the C4 query (0.75 against Grok-4.6 at 0.78) and the Sigma rule (0.90 against GPT-5.6-sol at 0.91). Because different models lead different stages (Tables~\ref{tab:stage_c1} to~\ref{tab:stage_c4}), we do not combine our metrics into a new overall reward; this suggests routing each stage to the model that performs it best as the natural next step.

\FloatBarrier   %
\section{Conclusion}
\label{sec:conclusion}

Open-weight models match the strongest closed models on this benchmark, and GLM-5.3-flash does
so for less than a tenth of what those closed models cost. GLM-5.3-flash reaches reward 0.74 at
\$2 a run against 0.71 for Grok-4.6 at \$28 and 0.69 for Claude-Opus-5 at \$50. The
condition is that the model is given enough turns: at the benchmark's published limit of 70 messages the
leading open-weight and closed models cannot be separated.

This dependence on the budget is itself a finding, not only a caveat. The published limit is hard, and a
truncated sample scores zero on checkpoints it had already passed, so a comparison there
partly measures which models terminate rather than which are capable. The limit constrains the
models that take more turns to reach an answer, and these are not necessarily the models that
answer worse: DeepSeek-v4-flash had 12 of its 50 samples truncated at 70 messages and none at
250, the budget from which its unconstrained result is carried. This is one reason the published
setting orders the leading models poorly.

The stage-level metrics of Section~\ref{sec:stages} add a second conclusion. No model leads
every stage: the best technique mapping, table choice, query testing, detection query and Sigma
rule come from six different models, three of them tied on query testing, and only GPT-5.6-sol leads two. A single overall reward ranks
models for a choice that the workflow does not require, and routing each stage to the model that
performs it best is the natural next step.

Five of the seven closed models have results carried from a lower budget at which no sample
reached the cap, so a higher budget could not have changed them. Each reported result rests on one measurement of each of the 50 tasks, so a
difference of a few hundredths between two models should not be read as an ordering. In
particular, the top of the ranking is shared rather than settled: GLM-5.3-flash's
run made entirely at the unconstrained budget would place it third, behind Kimi-K3 and Grok-4.6, by gaps whose
intervals include zero. The cost claim holds under either run: a
run at the unconstrained budget costs little on GLM-5.3-flash and a great deal on the most expensive
closed models. The claim therefore rests as much on cost as on reward.

\section*{Author Contributions}
Sahil Garg set up and ran the experiments, developed the evaluation metrics introduced in this
work, and wrote the paper. Ruhi Gottumukkala, as a summer research intern, carried out the error
analysis and related the performance of the open-weight models to their architectures and to
their scores on general benchmarks, to validate and interpret the results. Vishal Agarwal advised
on the benchmark as a cybersecurity expert and provided the resources that made the work
possible. Shubham Vishwakarma led the engineering that set up the benchmark for experiments at
scale.

\section*{Acknowledgements}
We thank Sridhar Dutta and Manish Varma Datla for their feedback on this benchmark.

\end{document}